\documentclass[conference]{IEEEtran}
\IEEEoverridecommandlockouts
\usepackage{cite}
\usepackage{amsmath,amssymb,amsfonts}
\usepackage{algorithm}
\usepackage{algorithmic}
\usepackage{graphicx}
\usepackage{textcomp}
\usepackage{xcolor}
\usepackage{tabularx}
\usepackage{multirow}

\begin{document}
% \title{Towards Fine-Grained Feature Analysis Bird Species Classification using Layer-wise Relevance Propagation\\}
\title{Analyzing the Impact of AI Tools on Student Study Habits and Academic Performance\\
}

\author{

\IEEEauthorblockN{Ben Ward}
\IEEEauthorblockA{
    \textit{Dept. of Computer Science} \\
    \textit{Kent State University} \\
    Kent, OH, USA \\
    bward29@kent.edu}

\and
\IEEEauthorblockN{Deepshikha Bhati}
\IEEEauthorblockA{
    \textit{Dept. of Computer Science} \\
    \textit{Kent State University} \\
    Kent, OH, USA \\
    dbhati@kent.edu}
\and
\IEEEauthorblockN{Fnu Neha}
\IEEEauthorblockA{
    \textit{Dept. of Computer Science} \\
    \textit{Kent State University} \\
    Kent, OH, USA \\
    neha@kent.edu}
\and

\IEEEauthorblockN{Angela Guercio}
\IEEEauthorblockA{
    \textit{Dept. of Computer Science} \\
    \textit{Kent State University} \\
    Kent, OH, USA \\
    aguercio@kent.edu}
\and

}

\maketitle

\begin{abstract}
This study explores the effectiveness of AI tools in enhancing student learning, specifically in improving study habits, time management, and feedback mechanisms. The research focuses on how AI tools can support personalized learning, adaptive test adjustments, and provide real-time classroom analysis. Student feedback revealed strong support for these features, and the study found a significant reduction in study hours alongside an increase in GPA, suggesting positive academic outcomes. Despite these benefits, challenges such as over-reliance on AI and difficulties in integrating AI with traditional teaching methods were also identified, emphasizing the need for AI tools to complement conventional educational strategies rather than replace them. Data were collected through a survey with a Likert scale and follow-up interviews, providing both quantitative and qualitative insights. The analysis involved descriptive statistics to summarize demographic data, AI usage patterns, and perceived effectiveness, as well as inferential statistics (T-tests, ANOVA) to examine the impact of demographic factors on AI adoption. Regression analysis identified predictors of AI adoption, and qualitative responses were thematically analyzed to understand students' perspectives on the future of AI in education. This mixed-methods approach provided a comprehensive view of AI’s role in education and highlighted the importance of privacy, transparency, and continuous refinement of AI features to maximize their educational benefits.

\end{abstract}


\section{Introduction}

Artificial Intelligence (AI) has emerged as a transformative force across various sectors, including education. Educational institutions are increasingly adopting AI-driven tools—such as adaptive learning platforms, AI tutoring systems, and smart note-taking applications—to deliver personalized learning experiences, streamline administrative tasks, and enhance student outcomes. By addressing individual needs, AI can improve study efficiency, reduce stress, and elevate academic performance. This aligns with the objectives of Education 4.0, which aims to prepare students for a technologically advanced future~\cite{ghimire2024generative}, \cite{elhussein2024shaping}.

While AI holds significant promise, its adoption in the education sector is influenced by various factors, including demographics, students' fields of study, and their comfort level with technology. Studies indicate that students in technical disciplines, such as computer science, are more likely to adopt AI tools and report positive impacts due to the alignment of AI capabilities with their academic needs. However, several barriers—such as privacy concerns, integration challenges, costs, and limited faculty support—continue to hinder the widespread adoption of AI tools in educational environments~\cite{ghimire2024generative}, \cite{elhussein2024shaping}, \cite{ cardona2023artificial}, \cite{akgun2022artificial}. These challenges highlight the complexity of AI integration and emphasize the need for ethical considerations regarding data privacy, equitable access, and maintaining a balance between AI-driven and human-centered learning ~\cite{ghimire2024generative}, \cite{elhussein2024shaping},\cite{eden2024integrating}.

This paper explores the impact of AI tools on students' study habits, specifically targeting academic performance, time management, and motivation. Utilizing a comprehensive survey from a variety of institutions, the study highlights both the advantages and limitations of AI in educational settings. It addresses key questions such as: How do AI-powered tools enhance academic outcomes? What factors influence their effective adoption? What barriers hinder their full integration? The findings offer valuable insights for educators, policymakers, and developers, emphasizing AI's crucial role in shaping the future of education while advocating for the responsible and inclusive implementation of these technologies~\cite{ghimire2024generative}, \cite{elhussein2024shaping}.

This paper makes the following key contributions to the field of AI in education:

\begin{itemize}
    \item \textbf{Comprehensive Analysis of AI’s Impact:} A holistic examination of AI's influence on students' study habits, covering time management, academic performance, and motivation across various disciplines.
    \item \textbf{Factors Influencing AI Adoption:} Identification of key demographic and contextual factors affecting AI adoption.
    \item \textbf{Barriers and Challenges:} Insight into obstacles, hindering AI adoption in education.
    \item \textbf{Framework for Future Research:} A framework for evaluating AI tools in education, balancing academic outcomes and student well-being.
    \item \textbf{Guidance for Educators and Policymakers:} Actionable recommendations for promoting responsible AI integration while safeguarding student privacy and autonomy.
\end{itemize}

\section{Related Work}

The application of AI in education has received considerable attention in recent years, with various studies analyzing its impact on learning processes, student engagement, and educational outcomes. Research on AI-driven educational tools has been particularly extensive, with studies often focusing on personalized learning systems, virtual tutors, and tools designed to support study efficiency and academic performance.

\subsection{AI for Personalized Learning and Student Engagement}
Recent studies have highlighted the potential of AI to personalize educational experiences by adapting content to individual learning styles and paces. The World Economic Forum’s report on Education 4.0 emphasizes AI's role in customizing learning pathways for students, helping them achieve more efficient outcomes through tailored instructional content and immediate feedback mechanisms \cite{hussin2018education}. Similarly, adaptive AI tools, such as intelligent tutoring systems and personalized learning platforms, have been shown to improve engagement and motivation, particularly among students in STEM fields \cite{lin2023artificial}.

AI has the potential to transform personalized learning and student engagement by tailoring educational experiences to individual needs and learning styles. Through data-driven insights, AI can provide real-time feedback on student performance, emotions, and engagement levels, enabling educators to customize teaching methods and interventions. This personalized approach not only enhances student motivation but also supports more effective learning outcomes, addressing the unique challenges faced by each learner in a sustainable education system \cite{lin2023artificial}.

Research demonstrates also the versatility of AI in enhancing engagement and support in specialized educational domains. For instance, AI-driven tools have been developed to aid caregivers of individuals with Autism Spectrum Disorder, focusing on the creation of personalized social stories to support learning and communication \cite{bhati2023bookmate} and integration of multimodal methods AI chatbot to provide tailored support \cite{francese2022multimodal}. Techniques further emphasize their role in fostering personalized learning experiences while addressing key challenges such as building student and educator trust, ensuring accessibility, and scaling solutions for diverse educational settings \cite{bhati2024survey}

\subsection{Comparative Studies on AI Tools and Traditional Methods}

Research comparing AI-driven study aids with traditional educational methods shows that AI tools, while enhancing academic performance, work best as complementary resources rather than replacements. AI systems assist students in organizing study routines, managing time, and providing immediate feedback—features that traditional methods often lack. However, successful integration of AI tools requires significant support from faculty, robust technology infrastructure, and structured training sessions to help students maximize the potential of these tools. Studies also emphasize that AI tools, such as intelligent tutoring systems and personalized learning aids, can significantly enhance educational experiences when used alongside traditional teaching methods, rather than as a substitute \cite{bahroun2023transforming}, \cite{alper2024evaluating}, \cite{asarsh2023ai}.

\subsection{Challenges and Ethical Considerations in AI Integration}
The integration of AI in education presents both significant opportunities and ethical challenges. Zawacki-Richter et al. (2019) \cite{martin2023systematic} provide a comprehensive review of AI applications, highlighting the challenges and ethical concerns associated with its use in education, including issues of fairness, bias, and transparency. Holmes et al. (2019) \cite{baker2019ai} further discuss the promises of AI in enhancing personalized learning and improving educational outcomes, while also addressing concerns about privacy, the potential for bias, and the evolving role of educators. Selwyn (2019) \cite{selwyn2019should} explores the ethical implications of AI potentially replacing human teachers, emphasizing the social and relational aspects of teaching that AI systems may not be able to replicate, raising questions about the dehumanization of education. These papers underscore the importance of balancing innovation with ethical considerations as AI continues to shape the future of education. While ethics is an important topic, this study primarily focuses on the effectiveness of AI tools in enhancing student learning. We are currently evaluating the ethical issues raised by these tools separately.

\section{Methodology}

\subsection{Participants}
\begin{figure}
    \centering
    \includegraphics[width=.5\linewidth]{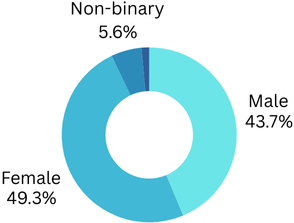}
    \caption{Gender distribution of
survey respondents}
    \label{fig:gender}
\end{figure}

\begin{figure}
    \centering
    \includegraphics[width=.4\linewidth]{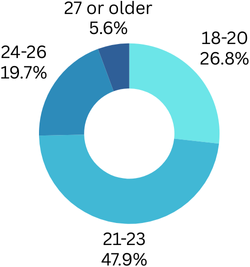}
    \caption{Age distribution of
survey respondents}
    \label{fig:Age}
\end{figure}

The study involved a diverse group of 71 university students (60 undergraduate and 11 graduate students) across various demographics, including age, gender, major, and academic level, collected through survey data. The age distribution of the survey respondents is plotted in Figure~\ref{fig:gender} and shows a nearly equal representation of male and female students, with few non-binary individuals and those who preferred not to disclose their gender. The age distribution is plotted in Figure ~\ref{fig:Age} and shows that the sample primarily included traditional college-aged students, with most between 21--23 years, followed by those aged 18--20, reflecting a typical undergraduate population with some graduate and non-traditional students. The academic level data of the respondents indicated that most participants were upper-level undergraduates (juniors and seniors), with a notable number of graduate students. The pie chart of academic majors in Figure~\ref{fig:grade_level} reveals a strong STEM representation, especially in computer science, with fewer students from humanities, healthcare, and other non-STEM fields.

\subsection{Survey Design}
The survey was structured to gather comprehensive data on AI tool usage, perceptions, and effectiveness across key sections:

\begin{itemize}
    \item \textbf{Demographics:} Collected information on age, gender, major, grade level, and university to analyze variations in AI usage based on these attributes.
    \item \textbf{AI Usage:} Focused on types and frequency of AI tool use (e.g., note-taking apps, AI tutors) and estimated percentage of study time involving AI.
    \item \textbf{Perceived Effectiveness:} Assessed AI’s impact on academic aspects like study efficiency and assignment completion using a Likert scale.
    \item \textbf{Comfort and Motivation Levels:} Rated participants' comfort with AI (scale of 1 to 5) and its effect on motivation, confidence, and engagement.
    \item \textbf{Challenges and Barriers:} Explored barriers to AI adoption, such as integration challenges, cost, and faculty support, with multiple response options.
    \item \textbf{Future Perspectives and Recommendations:} Collected qualitative data on students' views on AI’s future role in education and whether they would recommend it to peers.
\end{itemize}

\subsection{Analysis Methods}
Descriptive and inferential statistics were applied to analyze survey responses:

\begin{itemize}
    \item \textbf{Descriptive Statistics:} Calculated means, standard deviations, and frequency distributions for demographic data, AI usage patterns, and perceived effectiveness.
    \item \textbf{Inferential Statistics:} Used t-tests and ANOVA to examine differences in AI usage across demographic groups. Regression analysis identified predictors of AI adoption, with independent variables such as demographics, comfort level, and technology exposure.
    \item \textbf{Qualitative Analysis:} Conducted thematic analysis on responses in “Future Perspectives and Recommendations” to identify trends and insights on AI’s future role in education.
\end{itemize}

This mixed-methods approach enabled a comprehensive understanding of AI’s impact on education, combining quantitative analysis with qualitative insights.

\begin{figure}
    \centering
    \includegraphics[width=0.5\linewidth]{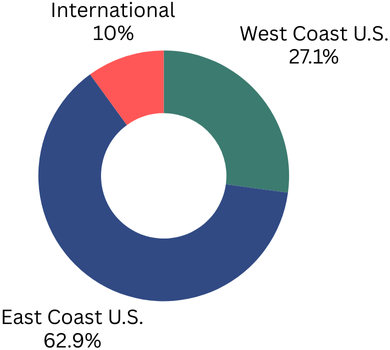}
    \caption{The distribution of survey respondents
across different universities}
    \label{fig:university}
\end{figure}

% This relatively balanced gender distribution helps validate that the survey findings represent perspectives across gender identities, though it remains within a predominantly STEM student population.

\section{Demographic Distribution of Survey Respondents}

According to Figure \ref{fig:gender}, which displays the age distribution of respondents, the majority were traditional undergraduate students. Of these, 47\% (35–40) were upper-level undergraduates aged 21–23, while 26\% (20) were freshmen or sophomores aged 18–20. Graduate students and older undergraduates aged 24–26 accounted for 19\% (15) of respondents, and only 5\% (5) were 27 years of age or older.

The age profile highlights a strong representation of the typical college age range (18–23), aligning with a higher concentration of junior and senior students. Although graduate and non-traditional students form a smaller proportion, they add diversity to the dataset. Figure \ref{fig:Age} complements this demographic composition, showing significant representation from technical universities. Together, these findings frame the survey’s insights within the context of early-20s undergraduate experiences and their usage of AI tools, while maintaining gender balance and institutional variety.

\section{Institutional Representation of Survey Respondents}

The geographical distribution in Figure 3 shows a strong U.S. coastal concentration, with East Coast universities representing 62.9\% of respondents and West Coast institutions at 27.1\%. International participants, comprising just 10\% of respondents, represented diverse global regions but with limited numbers from each area. Most international responses came from major research universities in Europe and Asia, with minimal representation from other continents. This U.S.-centric pattern (90\% total) reveals a significant regional imbalance. While the international responses provide valuable cross-cultural perspectives on AI tool adoption, their limited representation suggests the findings primarily reflect U.S. educational experiences and may not fully capture global variations in AI implementation, access, and cultural approaches to educational technology.

% The Pie chart shows the  survey respondents of geographical universities distributions about AI tools in education, as illustrated in Figure \ref{fig:university}:
% \begin{itemize}
%     \item The majority of responses (62.9\%) came from the East Coast U.S.
%     \item West Coast U.S. represented 27.1\% of respondents
%     \item International participants accounted for 10\%
% \end{itemize}

% \subsection{Key Observations}
% \begin{itemize}
%     \item Strong East Coast dominance with nearly two-thirds representation
%     \item Significant West Coast presence at over a quarter of responses
%     \item Limited international participation
%     \item Clear U.S.-centric response pattern (90\% total)
% \end{itemize}

% \subsection{Interpretation of Results}
% This distribution reveals:
% \begin{itemize}
%     \item Heavy concentration of responses from U.S. coastal regions
%     \item Notable regional imbalance favoring East Coast institutions
%     \item Limited global reach despite international participation
%     \item Potential geographical bias in survey findings
% \end{itemize}

% The distribution suggests that while the survey captured perspectives from multiple regions, there is a substantial skew toward U.S. coastal areas, particularly the East Coast, which may influence the overall representativeness of the findings.

\section{Grade Level Distribution}

\begin{figure}
    \centering
    \includegraphics[width=0.6\linewidth]{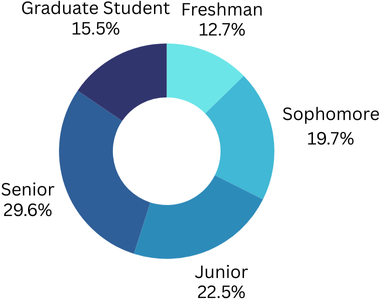}
    \caption{The distribution of survey respondents
by their academic grade level}
    \label{fig:grade_level}
\end{figure}

The donut pie chart in Figure \ref{fig:grade_level}) shows the academic distribution: Seniors (30\%, 21 students), Juniors (23\%, 16 students), Sophomores (20\%, 14 students), Graduate Students (15\%, 11 students), and Freshmen (13\%, 9 students). Upper-class students comprise the majority (53\%), suggesting higher AI tool engagement among more experienced students. While there's a general progression from lower to upper classes, the notable graduate student presence indicates AI adoption at advanced academic levels. The relatively low freshman participation may reflect less awareness or need for AI tools early in academic careers.

The survey's strong U.S. focus means findings primarily reflect American educational experiences and may not capture global variations in AI tool adoption and usage. Conclusions should be interpreted within the U.S. higher education context, with careful consideration before making broader international generalizations.

% This donut pie chart shows the distribution of survey respondents by their academic grade level. The largest group is Seniors at 30\% (21 students), followed by Juniors at 23\% (16 students), and Sophomores at 20\% (14 students). Freshmen make up 13\% (9 students) of respondents, while Graduate Students represent 15\% (11 students) as shown in Figure \ref{fig:grade_level}.

% \subsection{Key Observations}
% \begin{itemize}
%     \item Upper-class students (Juniors and Seniors combined) make up over half (53\%) of the respondents.
%     \item There's a fairly even progression from lower to upper classes, with numbers generally increasing by year.
%     \item Graduate students represent a significant minority.
%     \item Freshmen have the lowest representation among undergraduate levels.
% \end{itemize}

\section{Field of Study Distribution}

\begin{figure}
    \centering
    \includegraphics[width=.9\linewidth]{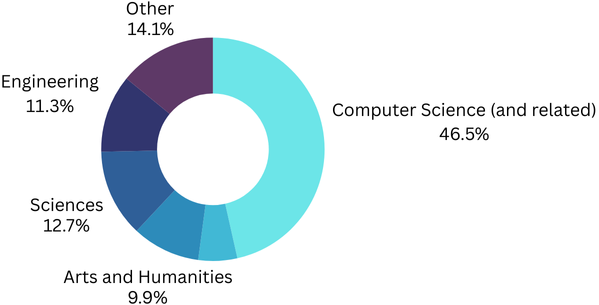}
    \caption{The chart displays the distribution of students who participated in the AI tools survey across different academic fields. }
    \label{fig:academic}
\end{figure}

The academic distribution in Figure \ref{fig:academic} shows Computer Science and related fields dominating at 46.5\%, followed by Other (14.1\%), Sciences (12.7\%), Engineering (11.3\%), Arts and Humanities (9.9\%), and Healthcare/Medicine (5.6\%). STEM fields collectively represent 70.5\% of respondents, suggesting higher AI tool engagement in technical disciplines. While the significant "Other" category indicates broader adoption, the lower representation in humanities and healthcare fields points to potential gaps in AI tool awareness or application in non-STEM areas. This distribution suggests opportunities for expanding AI tool adoption beyond technical fields while acknowledging the current concentration in STEM disciplines where immediate applications may be more apparent.

% This donut pie chart displays the distribution of students across different academic fields who participated in the survey about AI tools. The largest segment by far is Computer Science (and related fields) at 46.5\% of respondents, followed by Other at 14.1\%, Sciences at 12.7\%, and Engineering at 11.3\%. Arts and Humanities represent 9.9\% of respondents, while Healthcare and Medicine account for 5.6\% as shown in Figure \ref{fig:academic}.

% \subsection{Key Observations}
% \begin{itemize}
%     \item STEM fields collectively dominate the respondents (Computer Science, Engineering, and Sciences together make up 70.5\%).
%     \item Computer Science alone represents nearly half of all respondents.
%     \item Humanities and Healthcare fields have relatively lower representation.
%     \item The significant "Other" category suggests diverse adoption across various fields.
% \end{itemize}

% \subsection{Interpretation of Results}
% This distribution might reflect:
% \begin{itemize}
%     \item Greater awareness or accessibility of AI tools in technical fields.
%     \item More immediate applications of AI in STEM disciplines.
%     \item Possibly higher comfort levels with technology among STEM students.
%     \item Potential gaps in AI tool adoption or awareness in non-STEM fields.
% \end{itemize}

% The data suggests that while AI tools are being used across all academic disciplines, there's still a strong skew toward technical fields, indicating potential opportunities for broader adoption in the humanities and healthcare fields.

% Pie Chart 1: Perceived Academic Improvement
\section{Perceived Academic Improvement with AI Tools}
The bar chart in Figure 6 illustrates students' perceived academic improvement since they began using AI tools. The largest segment shows that 48\% (34 students) reported "Significant Improvement" in their academic performance. The second-largest group at 35\% (25 students) noted "Slight Improvement," while 17\% (12 students) reported "No Improvement".

The data reveals a strongly positive trend in the perceived academic impact of AI tools:
\begin{itemize}
    \item A substantial majority (83\%, combining both improvement categories) reported some level of academic improvement
    \item Nearly half of all students experienced significant improvement
    \item Only about one in six students reported no improvement
    \item The combined positive response suggests AI tools are effectively supporting academic performance
\end{itemize}

\begin{figure}
    \centering
    \includegraphics[width=0.6\linewidth]{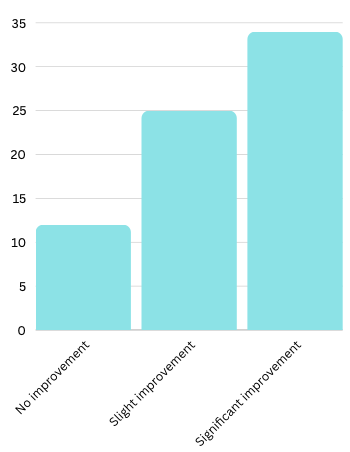}
    \caption{Perceived Academic Improvement Chart}
    \label{fig:Improvement}
\end{figure}

This distribution indicates that AI tools are generally having a positive impact on academic performance for most students who use them. The high percentage reporting significant improvement (48\%) is particularly noteworthy, suggesting that AI tools may be providing meaningful educational benefits. However, the presence of students reporting no improvement (17\%) indicates that AI tools may not be equally effective for all students or in all academic situations.

\vspace{1em}

% Pie Chart 2: Frequency of AI Tool Usage
\section{Frequency of AI Tool Usage for Studying}
\begin{figure}
    \centering
    \includegraphics[width=0.45\linewidth]{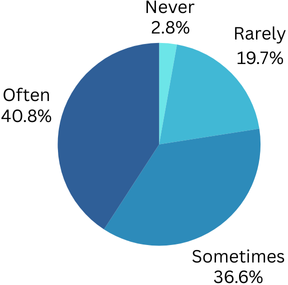}
    \caption{Pie chart displaying the frequency with which students use AI-powered study tools}
    \label{fig:AI-powered}
\end{figure}

The pie chart in Figure \ref{fig:AI-powered} displays the frequency with which students use AI-powered study tools. The largest segment shows that 41\% (29 students) use these tools "Often," followed by 37\% (26 students) who use them "Sometimes." A smaller portion, 20\% (14 students), reported using AI tools "Rarely," while only 2\% (2 students) indicated they "Never" use AI-powered study tools.

The data reveals a strong trend toward regular AI tool usage in studying:
\begin{itemize}
    \item The combined percentage of "Often" and "Sometimes" users (78\% or 55 students) shows that the vast majority of students are incorporating AI tools into their regular study routine
    \item Very few students completely avoid AI tools (only 2\%)
    \item A notable minority (20\%) are still hesitant or limited users
    \item The high percentage of frequent users suggests that students are finding value in AI-powered study tools
\end{itemize}

This distribution indicates that AI-powered study tools have become mainstream in student life, with most students using them at least sometimes. The very low percentage of non-users suggests that AI tools are becoming an integral part of modern study practices, though there's still variation in how frequently students choose to utilize them.

\vspace{1em}

% Pie Chart 3: Types of AI Learning Tools Used
\section{Types of AI Learning Tools Used}
\begin{figure}
    \centering
    \includegraphics[width=.9\linewidth]{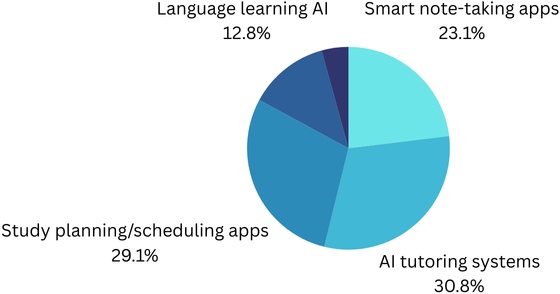}
   \caption{Pie chart showing the distribution of AI-powered learning tools currently used by students}
    \label{fig:used}
\end{figure}

The pie chart in Figure \ref{fig:used} chart shows the distribution of AI-powered learning tools currently used by students, where respondents could select multiple options. AI tutoring systems are the most popular, used by 36 students (31\% of responses), followed closely by study planning/scheduling apps with 34 users (29\%). Smart note-taking apps are used by 27 students (23\%), while language learning AI tools are utilized by 15 students (13\%). A small portion of students, 5 in total (4\%), indicated they use other AI-powered learning tools.

The distribution reveals several interesting patterns:
\begin{itemize}
    \item Students appear to value AI most for direct learning support (tutoring systems)
    \item Organization and planning tools are almost equally popular as tutoring systems
    \item Note-taking applications represent a significant but smaller portion of AI tool usage
    \item Language learning represents a specialized but notable use case
    \item The relatively even distribution among the top three categories suggests students are using multiple AI tools for different aspects of their studies
\end{itemize}

This data indicates that students are adopting AI tools across various aspects of their learning experience, with a particular emphasis on tools that provide direct educational support and help with organization. The relatively low percentage for "Other" suggests that most common AI learning tool needs are covered by the main categories listed.

\vspace{1em}

% Pie Chart 4: Study Time Spent Using AI Tools
\section{Study Time Spent Using AI Tools}
\begin{figure}
    \centering
    \includegraphics[width=0.4
    \linewidth]{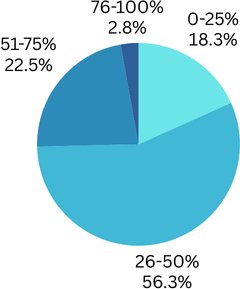}
    \caption{Pie chart illustrating the percentage of study time students spend using AI tools}
    \label{fig:time}
\end{figure}

This pie chart in Figure \ref{fig:time} illustrates the percentage of study time students spend using AI tools. The largest segment shows that 56\% of students (40 respondents) use AI tools for 26-50\% of their study time, representing a moderate level of AI integration. The next largest group at 23\% (16 students) uses AI tools for 51-75\% of their study time, while 18\% (13 students) use AI tools for 0-25\% of their study time. Only a small fraction, 3\% (2 students), reported using AI tools for 76-100\% of their study time.

This distribution suggests that most students are taking a balanced approach to incorporating AI into their studies:
\begin{itemize}
    \item The majority (56\%) use AI tools for roughly a quarter to half of their study time
    \item Very few students (3\%) are heavily dependent on AI tools
    \item A notable portion (18\%) uses AI tools sparingly
    \item About a quarter of students (23\%) use AI tools for more than half their study time
\end{itemize}

The data indicates that while AI tools have become a significant part of studying, most students are not overly reliant on them and maintain a mix of traditional and AI-assisted study methods. This balanced approach might represent a healthy integration of AI technology into academic work while maintaining traditional study practices.

\vspace{1em}

% Graph 1: Comfort Level with AI Technology
\section{Comfort Level with AI Technology}
\begin{figure}
    \centering
    \includegraphics[width=0.7\linewidth]{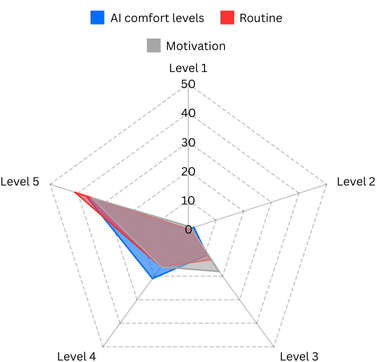}
    \caption{Radar chart displaying students' comfort levels with AI technology in dark blue color }
    \label{fig:comfort}
\end{figure}

The radar chart in Figure \ref{fig:comfort} displays in dark blue color students' comfort levels with AI technology on a 5-point scale. The results show a high average comfort level of 4.31 out of 5. Specifically, 37 students reported Level 5 (highest comfort), 21 students indicated Level 4, 11 students rated their comfort at Level 3, only 2 students reported Level 2, and notably, no students reported Level 1 (lowest comfort).

The data reveals that students generally feel very comfortable with AI technology, with 58 students (combining Level 4 and 5) reporting high comfort levels. This high comfort level could be attributed to:
\begin{itemize}
    \item Increasing exposure to AI tools in everyday life
    \item User-friendly interfaces of modern AI applications
    \item Growing up in a digital age
    \item Positive experiences with AI in educational settings
    \item Good institutional support for AI integration
\end{itemize}

The absence of any Level 1 rating and very few Level 2 ratings (only 2) suggests that severe discomfort with AI technology is rare among students. This high comfort level (4.31 average) indicates that most students are well-positioned to take advantage of AI tools in their educational journey, with very few feeling technologically intimidated.

\vspace{1em}

% Graph 2: Impact of AI on Study Routines
\section{Impact of AI on Study Routines}
The radar chart in Figure \ref{fig:comfort} displays in dark red color how students rate the AI's impact on their study routines, using a scale from 1 (Very Negative) to 5 (Very Positive). The results are notably positive with a high average rating of 4.37 out of 5. Looking at the distribution: 41 students gave the highest rating of Level 5 (Very Positive), 16 students rated it Level 4, 13 students gave a neutral Level 3 rating, only 1 student rated it Level 2, and notably, no students rated it Level 1 (Very Negative).

This overwhelmingly positive response (57 students rating Level 4 or 5) suggests that AI has significantly improved students' study routines. This could be attributed to several benefits:
\begin{itemize}
    \item More efficient research and information gathering
    \item Better organization of study materials
    \item Personalized learning assistance
    \item Flexible access to help when needed
    \item Streamlined note-taking and summarization capabilities
\end{itemize}

The absence of any Level 1 rating and only one Level 2 rating is particularly telling, indicating that even students who aren't strongly positive about AI's impact still don't find it detrimental to their study routines. The high average rating of 4.37 suggests that AI tools are successfully helping students establish more effective study habits and routines.

\vspace{1em}

% Graph 3: Impact of AI on Motivation
\section{Impact of AI on Motivation}
The same radar chart in Figure \ref{fig:comfort} illustrates in dark grey color how students rate the impact of AI tools on their motivation, using a scale from 1 (Very Negative) to 5 (Very Positive). The results show a notably positive trend with an average rating of 4.17 out of 5. The distribution breakdown shows that 35 students reported Level 5 (Very Positive), 16 students gave Level 4 ratings, 18 students indicated Level 3 (neutral), and only 1 student each rated Levels 2 and 1 (negative impacts).

The data strongly suggests that AI tools are generally having a positive effect on student motivation. The overwhelming majority of responses (51 students combining Level 4 and 5) indicate that students find AI tools motivating rather than demotivating. This could be due to several factors:
\begin{itemize}
    \item AI tools may make learning more engaging
    \item Instant feedback from AI tools can improve motivation
    \item Personalized learning experiences can make progress more visible and rewarding
    \item Students may feel more in control of their learning process
\end{itemize}

The very few negative responses (only 2 students) further reinforce the notion that AI tools are largely motivating students, helping them to stay engaged and focused on their academic work.

\section{User Experience and Feedback}

The feedback gathered from students highlights several areas where AI tools in education have a significant impact. Notably, many students expressed that AI tools are instrumental in saving time and enhancing understanding. Some of the key feedback is summarized below:

\subsection{Desired AI Features for Educational Tools}

\begin{itemize}
    \item \textbf{Adaptive Learning Path:} Several students suggested that AI should adapt the learning path in real-time based on student performance and feedback. This would enable a personalized learning experience.
    \item \textbf{Integration with Textbooks:} Some students expressed interest in seeing online textbooks integrate AI tools, allowing them to ask questions and get answers based on the contents of the book.
    \item \textbf{On-Demand AI Interaction:} A few students mentioned the desire to be able to easily push AI tools aside when not needed, indicating the importance of flexibility in engagement with AI.
    \item \textbf{Real-Time Classroom Feedback:} Students suggested that AI could enhance classroom interaction by analyzing expressions, movements, and language to assess students’ engagement and understanding.
    \item \textbf{Personalized Test Difficulty:} AI systems that adapt test difficulty based on student performance were also highly recommended to improve the learning experience.
    \item \textbf{Teaching Plan Generation and Optimization:} Several responses highlighted the value of AI helping teachers generate and optimize lesson plans tailored to the course syllabus and student needs.
\end{itemize}

\subsection{General Feedback on AI in Education}

Overall, feedback regarding AI in education has been positive. Students noted that AI tools greatly assist with tasks such as filling out study guides, offering explanations when unsure of a concept, and providing quick, personalized feedback. However, students also cautioned against over-reliance on AI and emphasized the need for good study habits.
Here are some of the key points extracted from the feedback.

\begin{itemize}
    \item \textbf{Enhanced Learning Efficiency:} Many students reported that AI tools contributed significantly to improving their learning efficiency.
    \item \textbf{AI as a Supplementary Tool:} Several students emphasized that AI should play a supplementary role in learning, with the real effect depending on the individual's learning efforts.
    \item \textbf{Collaboration Features:} Some students expressed interest in seeing more collaborative features integrated into AI tools, which could support group studies.
    \item \textbf{Data Security and Privacy Concerns:} Ensuring the security of student data and compliance with regulations such as GDPR or FERPA was highlighted as crucial by some students, with a focus on transparency regarding data usage.
\end{itemize}

\section{Conclusion}

This study highlights the promising role of AI in enhancing educational tools, with feedback from students showing strong support for features like adaptive learning paths, personalized test difficulty, and real-time classroom analysis. A significant drop in study hours and an increase in GPA demonstrate its effectiveness in improving academic performance. However, students emphasized the need for flexibility and transparency to ensure AI complements traditional teaching methods.

While AI tools contribute to efficient study routines and personalized learning, challenges like over-reliance and curriculum integration persist. Future advancements should refine these features, address privacy concerns, and promote collaboration, ensuring AI remains a supportive and ethical resource in education.

\bibliographystyle{IEEEtran}
\bibliography{main.bib}
% \vspace{12pt}

\end{document}